\documentclass[letterpaper, 10 pt, conference]{ieeeconf}  

\IEEEoverridecommandlockouts                              

\usepackage{multicol}
\usepackage{amsmath}
\usepackage{amssymb,amsfonts}
\usepackage{xfrac}
\usepackage{graphicx}
\usepackage[caption=false]{subfig}
\usepackage{mathtools}
\usepackage{algorithm}
\usepackage{algpseudocode}
\usepackage[dvipsnames]{xcolor}        
\usepackage{xspace}
\usepackage{multirow}
\usepackage{wrapfig} 
\usepackage{tabularx}
\usepackage{makecell}

\graphicspath{{./Figures}}
\usepackage{url}
\usepackage[%
hyperfigures=true,%
breaklinks=true,%
colorlinks=true,%
citecolor=BrickRed,%
linkcolor=OliveGreen,%
urlcolor=RoyalBlue%
]{hyperref}

\usepackage{cite}   

\usepackage{fancyhdr}
\newcolumntype{C}[1]{>{\centering\arraybackslash}p{#1}}

\usepackage{booktabs} 
\usepackage{siunitx} 

\usepackage{thmtools,thm-restate}

\newtheorem{problem}{Problem}

\def\figurename{Fig.}

\title{\LARGE \bf

Tunable Leg Stiffness in a Monopedal Hopper for Energy-Efficient Vertical Hopping Across Varying Ground Profiles
}

\author{Rongqian Chen, Jun Kwon, Kefan Wu and Wei-Hsi Chen
\thanks{R. Chen is at George Washington University (rongqianc@gwu.edu), K. Wu is at 
University of Connecticut (pwh24004@uconn.edu), J. Kwon and W.-H. Chen are with the General Robotics, Automation, Sensing \& Perception Lab at the University of Pennsylvania (\{junkwon, weicc\}@seas.upenn.edu)}
\thanks{We thank Cynthia Sung for her support on this project and Sonia Roberts for providing us with her Ground Emulator and her guidance in its modification and use. We also thank Rachel Holladay and Shivangi Misra for their valuable feedback and suggestions on the paper.}}%

\begin{document}

\maketitle
\thispagestyle{empty}
\pagestyle{empty}

\begin{abstract}

We present the design and implementation of HASTA (Hopper with Adjustable Stiffness for Terrain Adaption), a vertical hopping robot with real-time tunable leg stiffness, aimed at optimizing energy efficiency across various ground profiles (a pair of ground stiffness and damping conditions).
By adjusting leg stiffness, we aim to maximize apex hopping height, a key metric for energy-efficient vertical hopping.
We hypothesize that softer legs perform better on soft, damped ground by minimizing penetration and energy loss, while stiffer legs excel on hard, less damped ground by reducing limb deformation and energy dissipation. 
Through experimental tests and simulations, we find the best leg stiffness within our selection for each combination of ground stiffness and damping, enabling the robot to achieve maximum steady-state hopping height with a constant energy input. 
These results support our hypothesis that tunable stiffness improves energy-efficient locomotion in controlled experimental conditions.
In addition, the simulation provides insights that could aid in future development of controllers for selecting leg stiffness.
\end{abstract}

\section{Introduction}

While robotic locomotion can be optimized for energy efficiency on a single type of terrain, maintaining that efficiency across varying terrains poses a significant challenge.
In nature, humans~\cite{ferris1997interaction, farley1999leg} and animals~\cite{farley1993running, ferris1998running} dynamically adjust the stiffness of their limbs to adapt to changing ground conditions, enabling more energy-efficient movement across diverse surfaces. 
Inspired by this, researchers have explored tunable stiffness in robotics to improve adaptability and performance in diverse environments.


\subsubsection{Tunable Stiffness}
Tunable stiffness in robotics can be achieved through two primary methods: software-based virtual compliance and hardware-based variable stiffness actuators. 
Software-based virtual compliance, implemented via impedance control with force or positional feedback, allows real-time stiffness tuning through various control strategies, resulting in spring-like dynamics. Examples of this approach can be seen in robots like ANYmal~\cite{hutter2016anymal}, Minitaur~\cite{kenneally2016design}, and MIT Cheetah~\cite{bledt2018cheetah}. 
These systems benefit from fast response times due to the low-gearing of direct-drive motors~\cite{kenneally2020actuator, schumacher2019introductory}, but they also suffer from rapid heating due to joule losses~\cite{seok2014design}. 
Moreover, software-based solutions lack the energy storage benefits of passive compliance and require higher computational resources~\cite{fan2024review,kothanext}.

On the other hand, hardware-based tunable stiffness mechanisms use mechanical designs integrated with actuators to physically alter the system’s mechanical properties. 
These systems offer advantages such as reduced need for active energy input~\cite{ijspeert2014biorobotics, reher2021dynamic, 10161218}, enhanced stability and adaptability, and the ability to reduce high-impact forces~\cite{schumacher2019introductory, MOMBAUR2017135}. 
However, they tend to be larger and heavier due to their complex mechanisms and often have slower response times for real-time adjustments~\cite{vu2013knee}.

In this work, we introduce HASTA (Hopper with Adjustable Stiffness for Terrain Adaption), a vertical hopper with tunable stiffness achieved through a pneumatic bellows actuator developed in our previous work~\cite{chen2024design}. 
This system allows us to explore how stiffness adjustments in the compliant leg can enhance locomotion performance, offering energy storage benefits, improved dynamic response, and reduced weight and control complexity by eliminating the need for external air sources. 
In the context of vertical hopping, we aim to achieve energy-efficient locomotion, where energy efficiency is defined by the steady-state hopping height for a given fixed input energy.


\begin{figure}[tb]
    \centering
    \includegraphics[width = 0.95\linewidth]{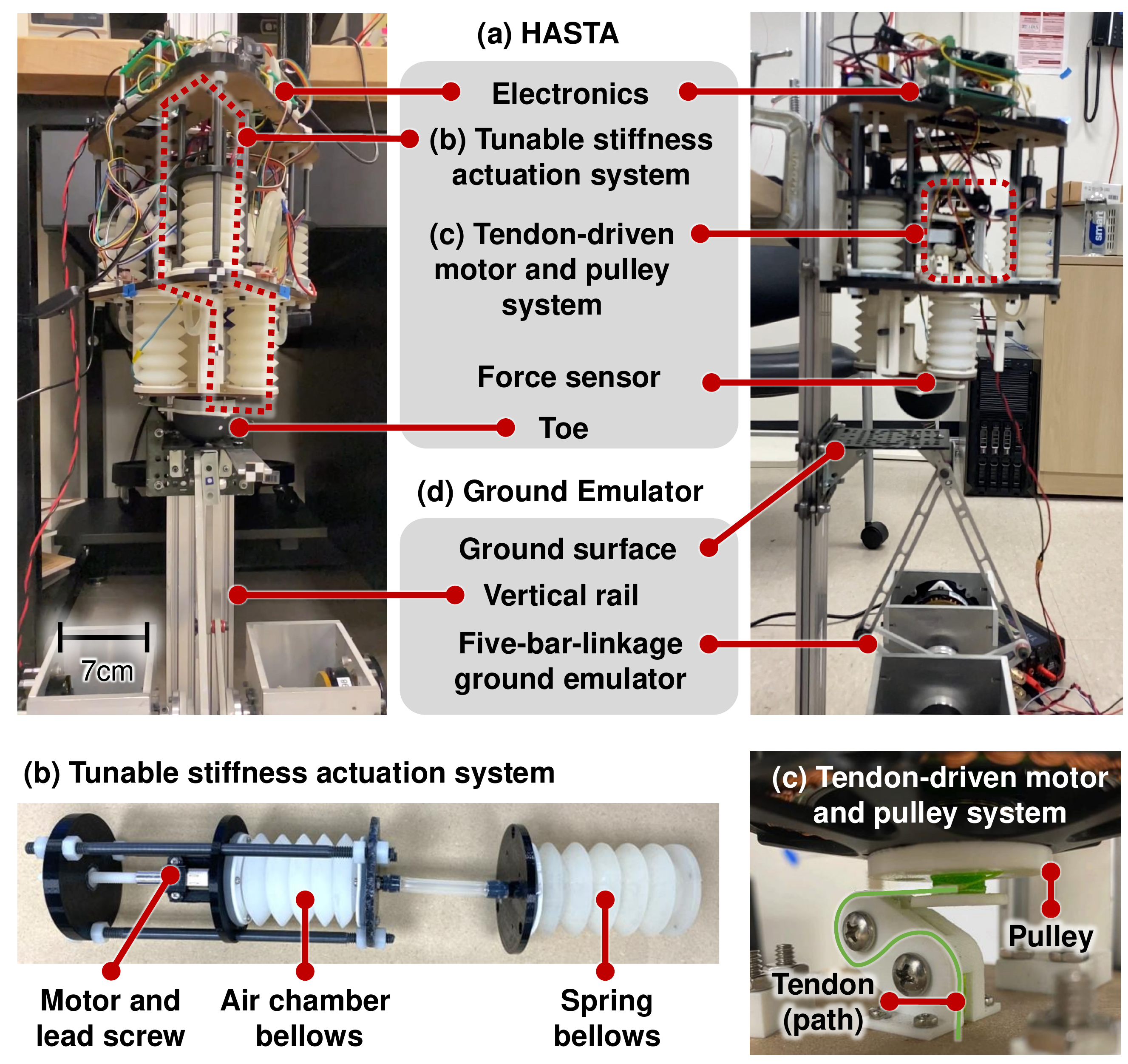}
    \vspace{-8pt}
    \caption{Anatomy of the (a) HASTA Robot, (b) tunable stiffness actuator, (c) tendon-driven actuator (highlighting the tendon), and (d) ground emulator.}
    \label{fig: mechanism}
    \vspace{-12pt}
\end{figure}

\subsubsection{Energy-efficient Locomotion on Various Ground Profiles}
Locomotion across diverse ground conditions requires balancing trade-offs between safety, robustness, workspace, and energy efficiency, making it a multi-objective optimization problem. 
As a first step, this work focuses on energy efficiency, investigating how tunable stiffness can optimize hopping performance across different ground profiles.

For analytical purposes, the ground is often modeled as a network of springs and dampers~\cite{vu2013knee, mutka2014adaptive, zhu2016compliance}. 
While research on tuning leg stiffness for different ground conditions exists, it has primarily focused on modeling and addressing the spring component of the ground~\cite{galloway2011experimental, vu2013knee, mutka2014adaptive, koco2016new, gurney2023uped}, with limited exploration of damping effects, particularly in real-world scenarios.
We argue that incorporating the damping component is essential for programming energy-efficient locomotion, as it influences energy dissipation and overall system performance.



Building on the work in~\cite{roberts2021virtual}, we further investigate energy-efficient vertical hopping on ground profiles with programmable stiffness and damping, with a focus on tuning stiffness in physical hardware to optimize energy efficiency. 
In this paper, we hypothesize that during vertical hopping, softer legs are more effective on soft, damped ground, minimizing penetration and energy loss, while stiffer legs perform better on hard, less damped ground by reducing limb deformation and energy dissipation. 
To our knowledge, this is the first study to systematically explore hardware-based tunable stiffness in a monopedal hopping robot across a range of both ground stiffness and damping conditions.

We validate our experiment through a simplified mass-spring-damper simulation, which demonstrates similar behavior to our experimental findings. 
Additionally, we explore how parameters such as leg damping and energy input affect performance, offering deeper insights into system behavior across different ground profiles. 
This simulation framework can also guide the selection of optimal stiffness settings in controllers for energy-efficient locomotion.


\subsubsection{Contribution and Organization}
In summary, our contributions are:
\begin{enumerate}
    \item Development of HASTA, a dynamic vertical hopper with real-time tunable stiffness for exploring various ground profiles. 
    It is equipped with one actuated degree of freedom (DOF) for tendon-driven actuation and three actuated DOFs for adjusting leg stiffness. 
    \item Real-world characterization of the relationship between leg stiffness and performance across different ground profiles.
    This characterization shows that as ground stiffness increases or damping decreases, stiffer legs reduce energy loss and achieve higher hopping heights. 
    \item Validation of a simulator for the described characterization, demonstrating that the simulation can be used to guide controllers in selecting optimal leg stiffness. 
\end{enumerate}

The paper is organized as follows: Sec.~\ref{secProblem} defines the problem statement and the robotic task. 
Sec.~\ref{secExperiment} introduces the robot platform and outlines the experimental procedure.
Sec.~\ref{secSimulation} describes the simulation setup and environment.
The results are presented and discussed in Sec.~\ref{secResult}.
Finally, Sec.~\ref{sec:conclusion} offers conclusions and directions for future work.

\section{Definition and Problem Statement}
\label{secProblem}

\begin{figure}[tb]
    \centering
    \vspace{6pt}
    \includegraphics[width = 0.85\linewidth]{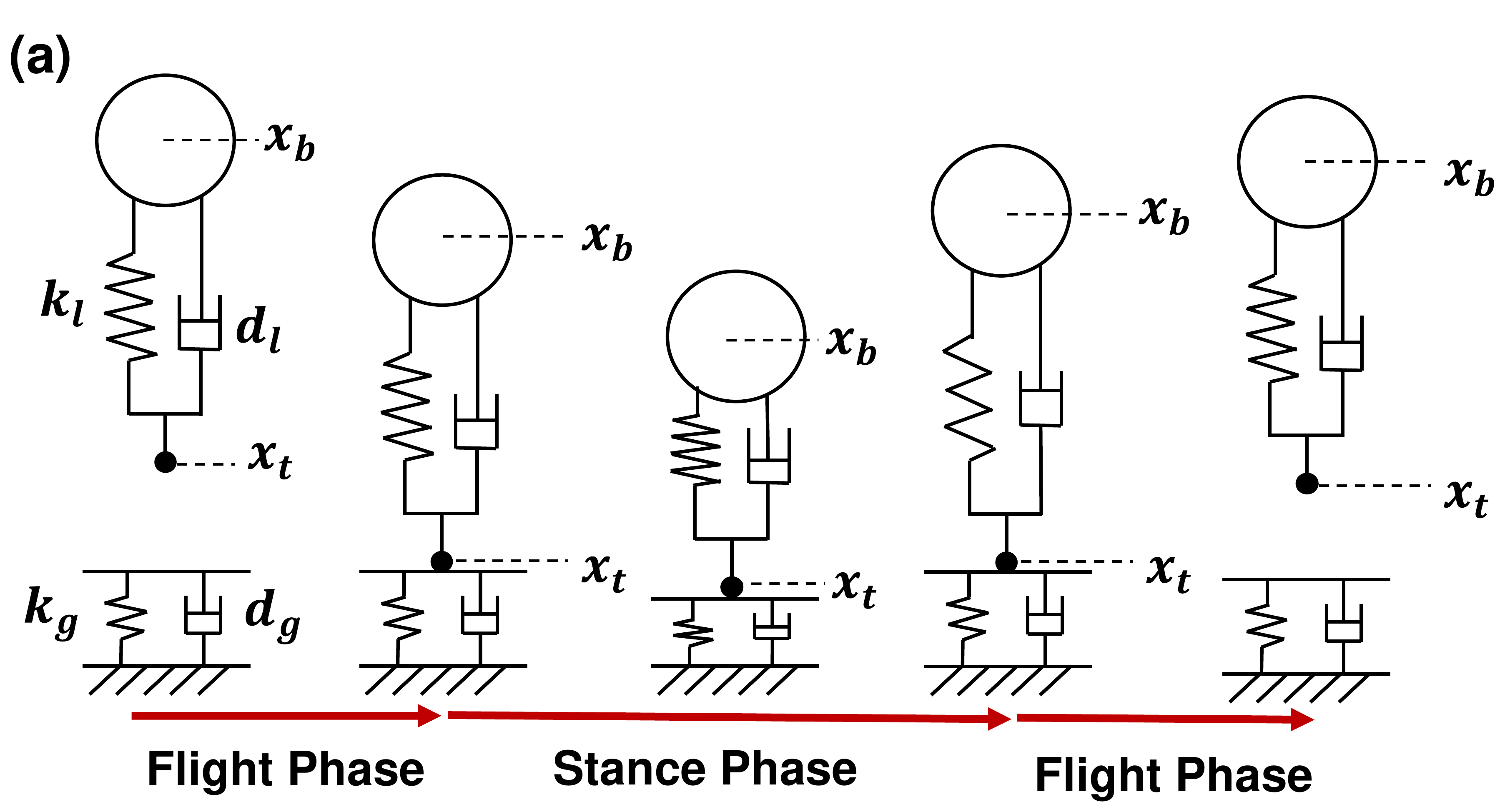}
    \includegraphics[width = 0.85\linewidth]{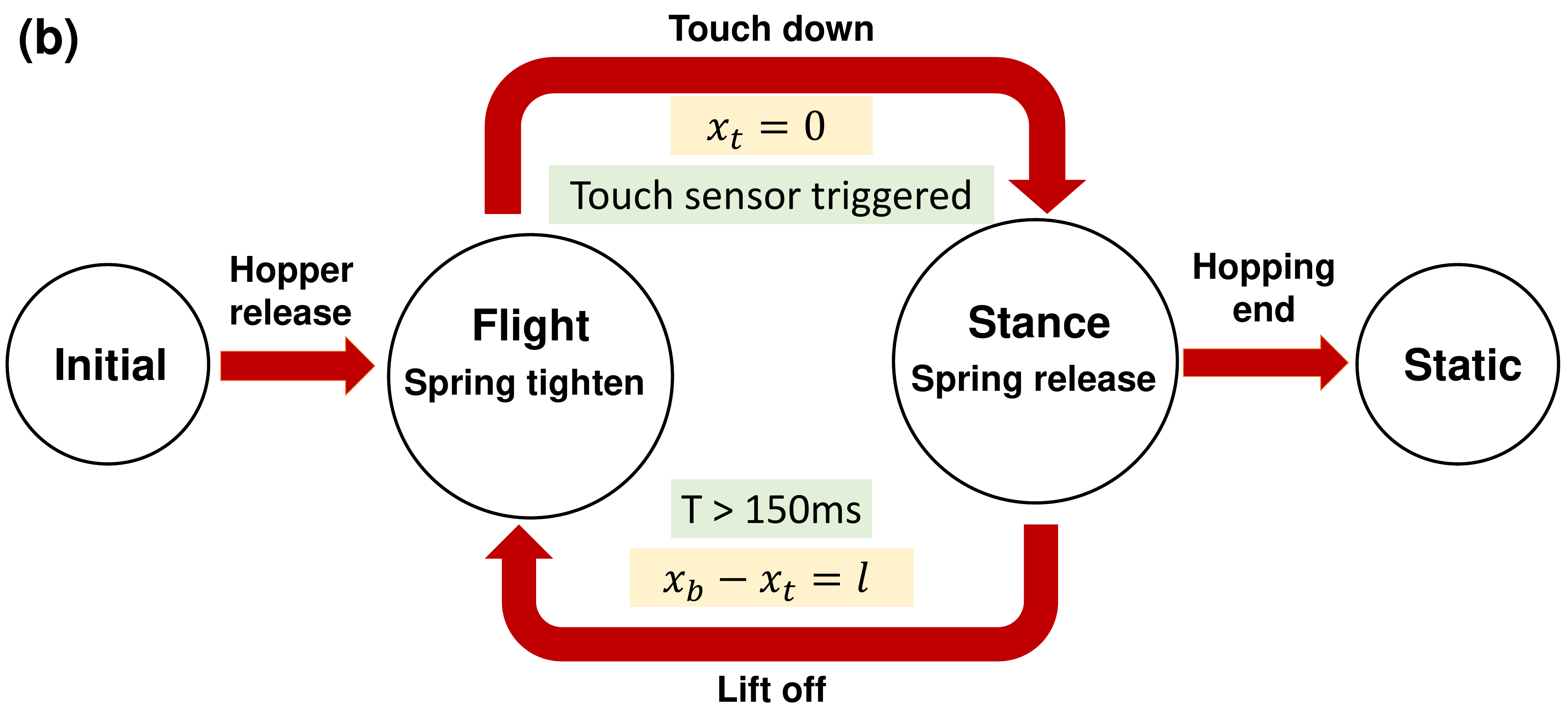}
    \vspace{-8pt}
    \caption{Task definition (a) still diagram of the hopper and ground emulator (b) state machine of the simulation.
    The touch-down and lift-off events conditions differ slightly in this paper due to physical constraints, where the yellow and green text boxes indicate the guards for the simulation and the experiments, respectively.}
    \label{fig: simulation}
    \vspace{-10pt}
\end{figure}



We aim to investigate how tunable stiffness in a robot's leg can be utilized for energy-efficient locomotion across varying terrains. 
In this study, we constrain the locomotion task to vertical hopping using a monopedal robot with a tunable stiffness leg, as this setup captures the fundamental interaction between leg stiffness and ground properties while isolating it from complexities such as gait dynamics and multi-legged coordination. 
Our goal is to demonstrate that an optimal leg stiffness setting exists, allowing the robot to achieve the maximum steady-state hopping height on different terrain types, given a constant energy input.
We argue that this steady-state hopping height is directly related to energy-efficient hopping.


The system of interest consists of a vertical hopper interacting with the ground (\figurename~\ref{fig: simulation} (a)). 
The \textit{vertical hopper} is constrained to vertical motion and consists of a body mass $m_b$ and a toe mass $m_t$, connected by a \textit{compliant leg} composed of a leg spring with tunable stiffness $k_l$ and an intrinsic damper with damping coefficient $d_l$.
The primary mode of operation involves storing and releasing energy in the compliant leg to propel the hopper upward, with tunable stiffness to manipulate energy storage.
The \textit{ground} is modeled as a flat surface on a Hookean spring with a spring constant $k_g$ and a damper with a damping coefficient $d_g$.
The \textit{ground profile }is defined by the pair $(k_g, d_g)$.


We define a \textit{vertical hopping task} using a state machine depicted in \figurename~\ref{fig: simulation}~(b), first introduced in~\cite{chen2020tendon}.
During the \textit{flight phase}, the leg spring is compressed to a controlled deformation $p$, storing potential energy. 
When the hopper \textit{touches down}, i.e., when the toe makes contact with the ground, the hopper transitions to the stance phase. 
In the \textit{stance phase}, the pre-compressed spring extends back to its original rest length, releasing energy into the ground and propelling the hopper upward. 
Once the toe leaves the ground, or \textit{lifts off}, the hopper re-enters the flight phase. 
We define the steady-state apex hopping height $h_{apex}$ as the maximum vertical displacement reached by a vertical hopper after the hopping height has stabilized.

We aim to demonstrate that for a given amount of energy input — provided as potential energy $E_{in}$ — there exists an optimal leg stiffness setting that allows the hopper to achieve the maximum jump height on various terrain types, or 
\begin{problem}
\label{prob1}
Given a vertical hopper, a constant input potential energy $E_{in}$ for each hop, 
a set of achievable leg stiffness values $\{k_l\}$, 
and a set of ground profiles $\{(k_g, d_g)\}$, 
find the leg stiffness from $\{k_l\}$ that maximizes the steady-state apex height $h_{apex}$ for each ground profile in $\{(k_g, d_g)\}$.
\end{problem}

We will address this problem experimentally through our designed vertical hopper, called \textit{HASTA} (Hopper with Adjustable Stiffness for Terrain Adaption). 
HASTA is capable of hopping using its compliant leg, which features tunable stiffness.
Additionally, we construct a simulation to validate our experimental findings and explore optimal leg stiffness configurations under different conditions.
Finally, we propose a strategy to enable energy-efficient hopping for a vertical hopper by selecting the leg stiffness that maximizes the apex height for each ground profile.

\section{Experiment}
\label{secExperiment}
We present the design and implementation details of the robot and ground emulator, followed by a description of the experimental setup and procedure to validate Problem~\ref{prob1}.

\subsection{Robot Platform: HASTA}
\begin{figure}[t]
    \centering
    \vspace{6pt}
    \includegraphics[width = 0.95\linewidth]{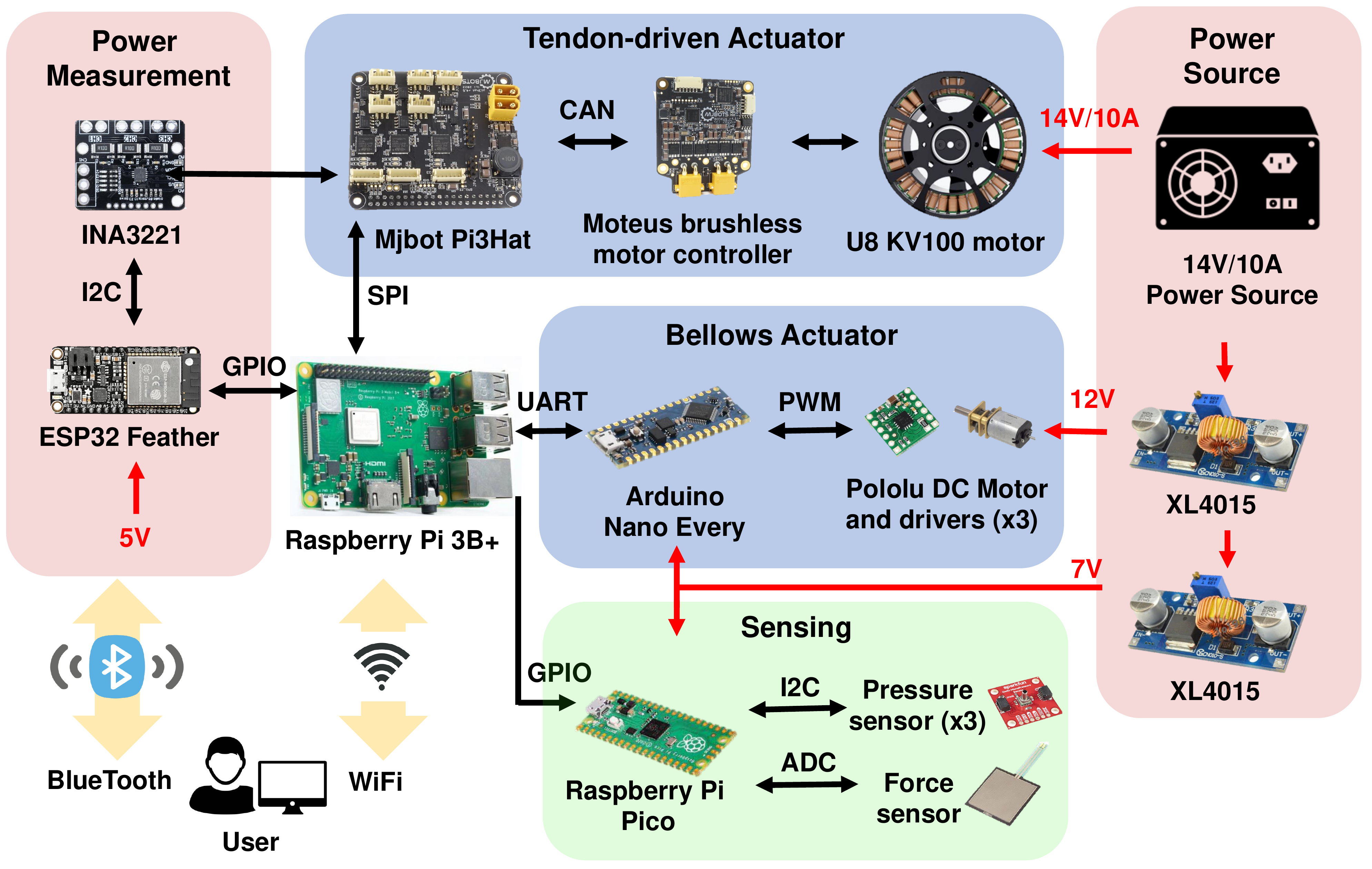}
    \vspace{-8pt}
    \caption{Infrastructure of HASTA}
    \label{fig: infrastructure}
    \vspace{-18pt}
\end{figure}

HASTA (Hopper with Adjustable Stiffness for Terrain Adaption) is a vertical hopper that can change its leg stiffness during execution time, as shown in \figurename~\ref{fig: mechanism}.
The design is a modification of the REBO Hopper~\cite{chen2020tendon} with an additional tunable stiffness actuation.
The hopper robot is mounted on a vertical rail to constrain its motion to vertical hopping.
HASTA comprises four main subsystems:
(a) the tunable stiffness system, (b) the tendon actuation system, and (c) the mechatronic system. 
The robot's specifications are listed in Table~\ref{tabHASTA} and the infrastructure are shown in \figurename~\ref{fig: infrastructure}.



\begin{table}[b]
    \scriptsize
    \vspace{-12pt}
    \caption{Specifications and Parameters}
    \label{tabHASTA}       
    \vspace{-12pt}
    \begin{center}
    \begin{tabular}{l l}
    \hline\noalign{\smallskip}
     Component and symbol& Properties\\\noalign{\smallskip}\hline\noalign{\smallskip}
     \textbf{HASTA specification} &\\
     Leg spring rest length, $l$ & \SI{97.5}{mm}\\
     Body mass (without toe), $m_b$ & \SI{2.5}{\kilo\gram}\\
     Toe mass, $m_t$ & \SI{0.3}{\kilo\gram}\\
     \noalign{\smallskip}\hline\noalign{\smallskip}
     \textbf{Experiment parameters} &\\
    Leg Stiffness, $k_l$           & 3351, 4279, 5341  (\SI{}{N/m})\\ 
    Ground Stiffness, $k_g$        & 2401.7, 3410.8, 4420  (\SI{}{N/m}) \\ 
    Ground Damping Coefficient, $d_g$ & 17.1, 35.2, 53.3, 71.4  (\SI{}{Ns/m})\\
    Energy Input, $E_{in}$ & 0.97 (\SI{}{J})\\
    \noalign{\smallskip}\hline\noalign{\smallskip}
    \textbf{Simulation parameters} &\\
    Leg Stiffness, $k_l$             & 3000, 4000, 5000 (\SI{}{N/m})\\
    Leg Damping Coefficient, $d_l$ & 30, 35, 40 (\SI{}{Ns/m})\\
    Ground Stiffness, $k_g$          & [2400 : 200 : 5400]* (\SI{}{N/m})\\
    Ground Damping Coefficient, $d_g$ & [15: 5 : 75]* (\SI{}{Ns/m})\\
    Input Energy, $E_{in}$          & 1, 1.56, 2.25 (\SI{}{J})\\ 
    \noalign{\smallskip}\hline
    \end{tabular}
     \vspace{3pt}
     \\$^*$[A:B:C] represents the value range from A to C with the increment of B
    \end{center}
\vspace{-8pt}
\end{table}


The tunable stiffness system consists of three \textit{pneumatic bellows actuators} that adjust stiffness by a factor of 1.43 without requiring an external air source. 
Each actuator comprises two bellows: the \textit{spring bellows}, whose stiffness varies with pressure, and the \textit{air chamber bellows}, which regulates pressure through deformation.
Further details on stiffness control and related works can be found in~\cite{chen2024design}. 

The tendon actuation system uses a brushless DC motor (T-Motor U8 KV100) to drive a UHMWPE tendon (Emma Kites) through the compliant leg.
One end of the tendon is attached to a motor-mounted pulley (\SI{7.5}{mm} radius, 3D-printed carbon fiber) and the other to a \SI{2.5}{in} semi-hemispherical silicone toe. 
Due to the leg's compliance and varying tendon spool radius, leg compression $p$ depends on both the motor angle $\theta_m$ and leg stiffness $k_l$. A lookup table $p = f_p(\theta_m, k_l)$ was experimentally generated to relate pre-compression $p$ to $\theta_m$ and $k_l$.
A proportional-derivative (PD) feedback control loop is implemented on the motor for position control with encoder data. 
The controller is able to maintain compression with an error less than $2\%$ in all tests for leg stiffness ranging from \SI{3351}{N/m} to \SI{5341}{N/m}, and pre-compression up to \SI{23}{mm}.
A silicone-molded force detection pad with a touch sensor for quick ground detection is placed between the toe and leg. The aluminum housing ensures both structural integrity and light weight.

The mechatronic system consists of four microcontrollers.
The primary controller is a Raspberry Pi 3B+, which controls the tendon-driven motor via the mjbots pi3hat and motor control board (https://mjbots.com/), and transmits collected data to the user via Wifi. 
The system includes three subsystems:
An Arduino Nano Every, which controls three DC motors in the tunable stiffness actuation system; 
A Raspberry Pi Pico, responsible for reading high-frequency sensory feedback, including air pressure in the tunable stiffness actuator and voltage from the touch sensor;
An ESP32 Feather, used to measure current draw and voltage from the robot, and send feedback data to the user via Bluetooth.
The system is powered by a 14V/10A supply, with voltage regulated for different subsystems including the computing units, the sensors, and the actuators via power breakout boards.


\subsection{Ground Emulator}
\label{sec:emulator}

\begin{figure}[tb]
    \centering
    \vspace{6pt}
    \includegraphics[width = 0.5\linewidth]{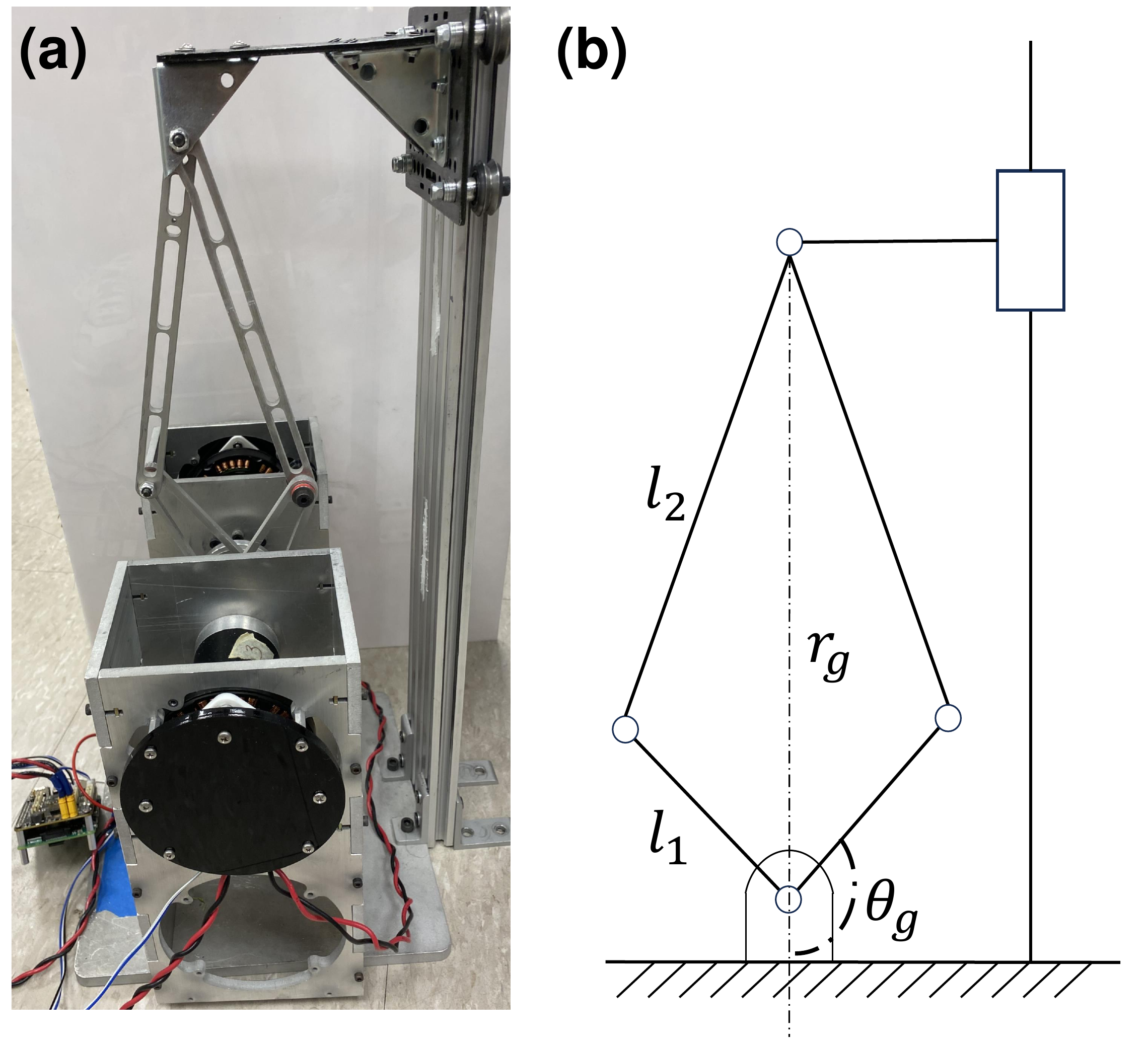}
    \vspace{-8pt}
    \caption{Ground emulator (a) photo (b) schematic drawing}
    \label{fig: ground emulator}
    \vspace{-12pt}
\end{figure}

We use a ground emulator, previously developed by S. Roberts and detailed in~\cite{roberts2019mitigating}, to precisely and systematically control ground properties, replicating a surface modeled as a parallel spring-damper system. 
The ground emulator uses a five-bar-linkage mechanism with link lengths $l_1$ and $l_2$, as depicted in \figurename~\ref{fig: ground emulator}. 
The ground surface is mounted on a linear rail to constrain it to vertical motion, and the height $r_g$ is controlled by two motors' angle $\theta_g$. 
A Raspberry Pi 3B+ is deployed for motor control and can be run remotely by the user through a Wi-Fi connection. 

The forward kinematics of the system mapping the motor angle \( \theta_g \) to the ground surface position \( r_g \) can be written as
\begin{equation}
r_g = -l_1 \cos \theta_g + \sqrt{-\tfrac{l_1^2}{2} + l_2^2 + \tfrac{l_1^2}{2} \cos 2\theta_g}.
\end{equation}
The required motor torque \( \tau_m \) is then computed as:
\begin{equation}
\tau_m = F \cdot \tfrac{dr_g}{d\theta_g},
\end{equation}
where $F = K_p \Delta r_g + K_d \dot{r}_g$ is the control force based on a proportional-derivative (PD) controller with the proportional ($K_p$) and derivative ($K_d$) gains, \( \Delta r_g \) is the position error, and \( {dr_g}/{d\theta_g} \) is the Jacobian.

To characterize the actual stiffness and damping coefficients of the ground emulator, we designed an experiment to map the controller's proportional gain (P gain, $K_p$) and derivative gain (D gain, $K_d$) to the corresponding equivalent stiffness and damping values.
We measured the oscillating positional trajectories of the ground surface under 3 different weights $(m_w)$, across 5 different $K_p$ and 5 different $K_d$ values.
The data were fitted to an underdamped mass-spring-damping model to determine the emulated stiffness and damping ratio.
A marker on the weight’s center was tracked using “Tracker” (https://physlets.org/tracker/).
In each trial, the weight was manually pushed down and released to oscillate until they stopped. 
Each trial was repeated three times, and the trajectory was fitted to the underdamped harmonic oscillator equation:
\begin{equation}
r(t) = A \exp(-\beta t) \cos\left(\sqrt{\alpha - \beta^2} \, t + \phi\right) - \tfrac{g}{\alpha},
\end{equation}
where \( A \) represents the oscillation amplitude, \( \alpha={k_g}/{m_w} \) and \( \beta={d_g}/{(2m_w)} \) relate back to the system's stiffness and damping, and \( \phi \) is the phase angle. 
The term \( g/\alpha \) accounts for the gravitational effect on the system.
The segmented trajectory data was fit to this model using MATLAB's \texttt{fit} function, and the parameters \( A \), \( \alpha \), \( \beta \), and \( \phi \) were estimated, allowing us to calculate $k_g$ and $k_l$.
The $R^2$ value of the fit for the entire experiments is above 0.9, indicating the accuracy of the estimation.
We then performed a linear fit across all the 75 data sets and obtained the two mappings: $k_g=f_k(K_p)$ and $d_g=f_d(K_d)$. 

\subsection{Experimental Methodology}
\label{sec: experimental procedure}


We aim to demonstrate that, with a constant energy input, the robot achieves different steady-state hopping heights for various programmed leg stiffness settings across different ground profiles $(k_g,d_g)$.
To account for energy loss during hopping, we pre-compress the leg spring by a distance $p = \sqrt{(2E_{in}/k_l)}$ to inject energy into the hopper.
We set the input energy $E_{in}=\SI{0.97}{J}$, and this value remains constant throughout all tests. 

\subsubsection{State machine}
The same state machine in \figurename~\ref{fig: simulation} with slightly different guard conditions controls the HASTA robot.
Starting with the initial state, the robot enters the flight phase after detecting a free fall and compresses its leg. 
Upon ground impact, sensed by a toe-embedded touch sensor, it transitions to the stance phase, where its leg tries to return to the uncompressed state. 
Due to the difficulty of detecting lift-off, a fixed time-based trigger of \SI{150}{ms} is used to transition the robot into the flight phase.
Camera data confirm that this duration allows full leg extension and energy transfer. 
Data logging begins after three impacts to ensure a steady state and stops after a preset number of impacts.

\subsubsection{Procedure}

Control gains of the ground emulator were selected using the lookup table from Sec.~\ref{sec:emulator} to achieve the desired stiffness and damping.
The selected parameters encompass properties of common surfaces, such as rubber and foam.
Leg stiffness were controlled following the method in~\cite{chen2024design} with parameters from Table~\ref{tabHASTA}. 
While we cannot control or measure the leg's damping properties, our simulations confirm that trends remain consistent despite this limitation.
Three visual markers tracked the robot (ground, body, leg), with an LED indicator syncing visual and sensor data.
A \SI{240}{Hz} slow-motion camera recorded the process, and the robot was manually released for hopping. 
After three hops, data collection started, capturing power, motor feedback, and sensor data. Post-test, video recordings were calibrated in MATLAB, and trajectories were extracted with Tracker.


\section{Simulation}
\label{secSimulation}

We construct a simulation of a vertical hopper hopping with different leg stiffness and various ground parameters, based on the state machine in \figurename~\ref{fig: simulation}. 
Given a constant energy input, we investigated the apex height of the steady-state hopping motion as a performance metric to evaluate the system's efficiency. 
Additionally, we examined how the apex height changes with variations in leg stiffness, damping coefficients, and input energy.

The system has two states, and the dynamics of the system during the flight phase can be expressed as follows:
\begin{equation}
\label{eq:flight}
    \begin{split}
m_b \ddot{x}_b &= F_l - m_b g \\
m_f \ddot{x}_f &= -F_l - m_f g,
\end{split}
\end{equation}
where $F_l = k_l (l - p - x_b + x_t) - d_l (\dot{x}_b - \dot{x}_f)$ is the internal force of the leg.
The transition from the flight phase to the stance phase occurs when the ground impact is detected, specifically when the toe position satisfies $x_t = 0$, indicating contact between the toe and the ground. 

During the stance phase, the hopper leg releases its pre-compression, and this phase concludes once the leg returns to its original rest length. 
The dynamics of the system during the stance phase can be expressed as follows:
\begin{equation}
\label{eq:stance}
    \begin{split}
m_b \ddot{x}_b &= F_l - m_b g \\
m_f \ddot{x}_f &= -F_l + F_g - m_f g,
\end{split}
\end{equation}
$F_l = k_l (l - x_b + x_t) - d_l (\dot{x}_b - \dot{x}_f)$ is the internal leg force and $
F_g = -k_g x_t - d_g \dot{x}_f$ is the ground supporting force. 
The stance phase ends when leg length is restored to $l$. 

To evaluate energy efficiency across different leg stiffness and ground profiles, we perform a grid search over 208 different ground profiles and three leg stiffness values.
Using MATLAB's \texttt{ODE45}, the simulation (based on Eq.~\eqref{eq:flight} and~\eqref{eq:stance} with Table~\ref{tabHASTA} parameters) ran until the preset hopping duration was reached, outputting body and toe trajectories and the steady-state apex height. 
This height, averaged over the last 10 hops, had a variance below \SI{0.001}{mm}. 
Trials where the robot failed to lift-off were marked as failures.

\section{Result}
\label{secResult}
We present the results of both the simulation and the experiment and discuss how the experimental data aligns with the simulation findings to validate our hypothesis.

\subsubsection{Experimental Result}

Following the experimental procedures outlined in Section~\ref{sec: experimental procedure}, we conducted experiments to evaluate hopping performance at different leg stiffness levels.
With constant potential energy applied during the hopper's flight phase, we measured the steady-state apex height using the parameters detailed in Table~\ref{tabHASTA}.

\begin{figure}[tb]
    \centering
    \vspace{8pt}
    \includegraphics[width = 0.81\linewidth]{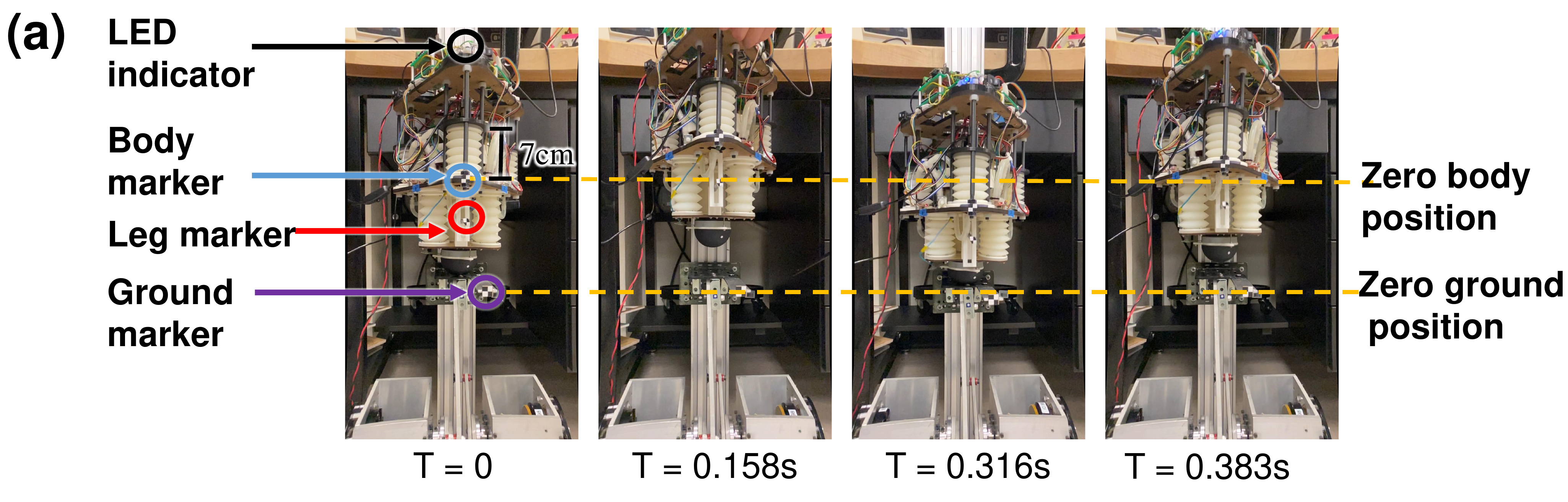}
    \vspace{-12pt}
    \includegraphics[width = 0.8\linewidth]{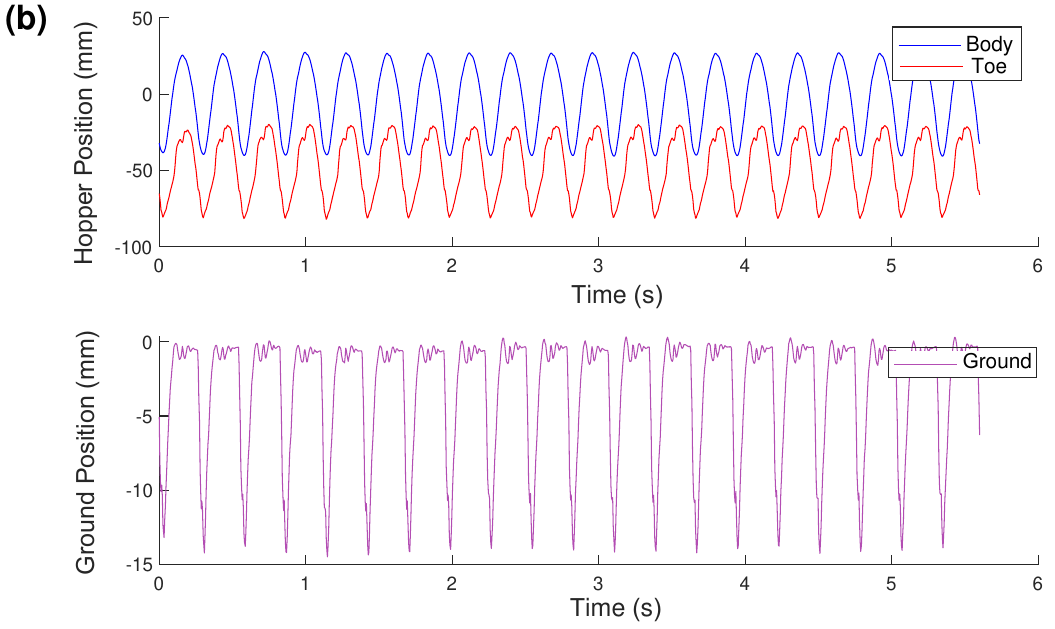}
    \vspace{-0pt}
    \caption{Hopping trajectory of an example trail where $k_g=\SI{4420}{N/m}$, $d_g=\SI{35.2}{Ns/m}$, $k_l=\SI{4280}{N/m}$  (a) photo shots of a trail (b) the trajectories of three markers.}
    \label{fig: hopping result}
    \vspace{-12pt}
\end{figure}

\figurename~\ref{fig: hopping result}~(a) shows still shots of the HASTA robot hopping on the ground emulator, while \figurename~\ref{fig: hopping result}~(b) presents its trajectory after three jumps, once steady-state was reached.
The zero position on both subplots represents the rest positions of the robot and the ground surface.
The blue and red trajectories correspond to the positions of the body and the toe, respectively. 
Due to its lighter weight, the toe exhibits vibrations during flight, and similar vibrations occur in the ground surface as it returns to rest.
 
\begin{figure}[t]
    \centering
    \includegraphics[width = 0.8\linewidth]{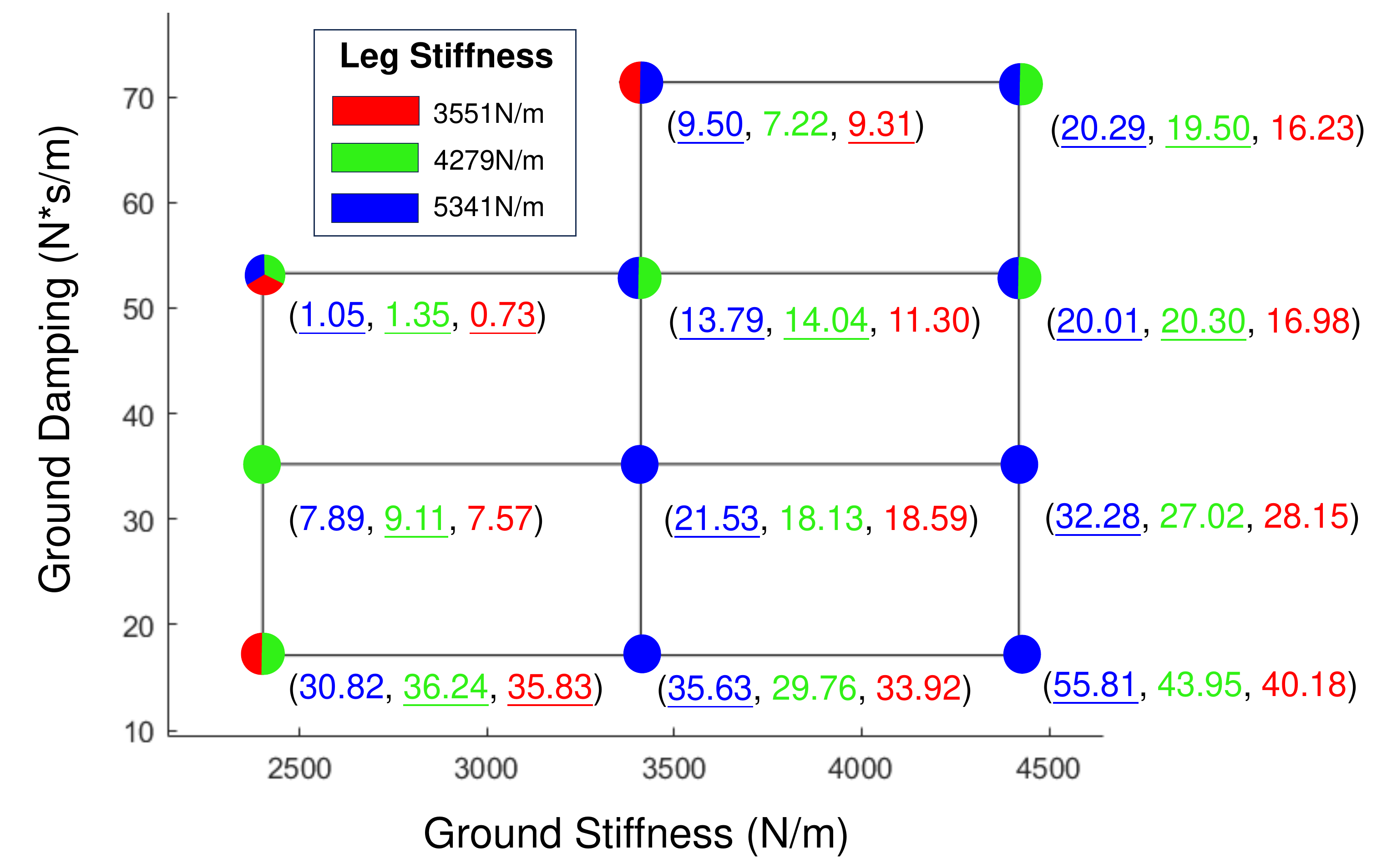}
    \vspace{-8pt}
    \caption{Experimental results of apex hopping height. The parenthesized values indicate the average apex height (mm), color-coded by leg stiffness.}
    \label{fig: exp height figure}
    \vspace{-12pt}
\end{figure}

The results of the apex height search experiment are shown in \figurename~\ref{fig: exp height figure}. 
We use the average apex height of 20 jumps as the performance metric.
The steady-state apex heights for different leg stiffness values across various ground profiles are displayed, with each value having a standard deviation of less than \SI{2}{mm}.
We used color coding to differentiate between the various leg stiffness.
For each ground profile $(k_g, d_g)$,
we highlighted the leg stiffness that resulted in the maximum steady-state hopping height by coloring the corresponding data points and underlining the value. 
In cases where the difference in apex heights between trials with different leg stiffness values was less than \SI{1}{mm} (below the standard deviation), we considered them too close to distinguish and treated them as ties. 
In such cases, the data point is split into multiple colors, indicating which leg stiffness values resulted in the maximum apex height.
We discarded data for the ground profile $(\SI{2401.7}{N/m}, \SI{71.1}{Ns/m})$ as the stiffer leg ($k_l=\SI{5341}{N/m}$) was unable to hop in this region. 

\begin{figure*}[tb]
     \centering
     {\includegraphics[width = 0.24\linewidth]{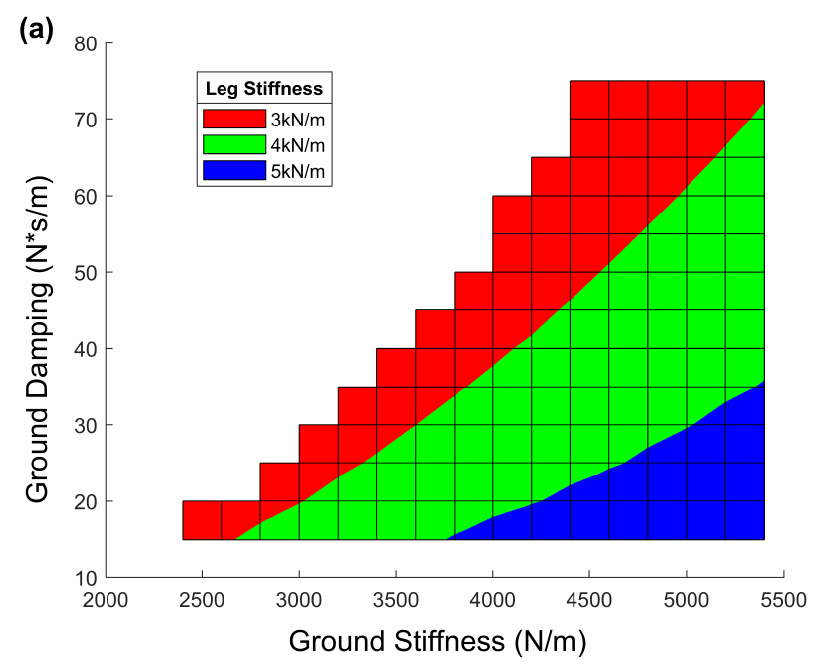}}
     {\includegraphics[width = 0.24\linewidth]{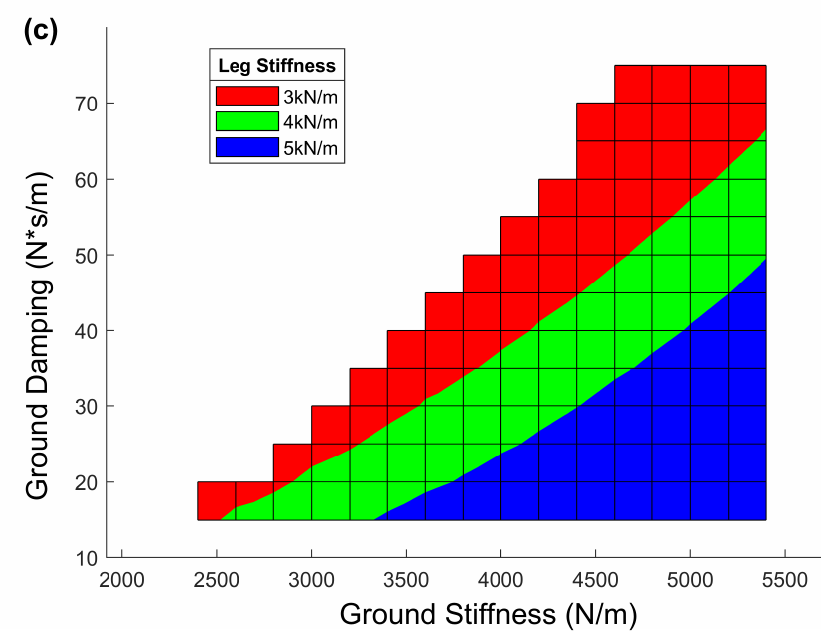}}
     {\includegraphics[width = 0.24\linewidth]{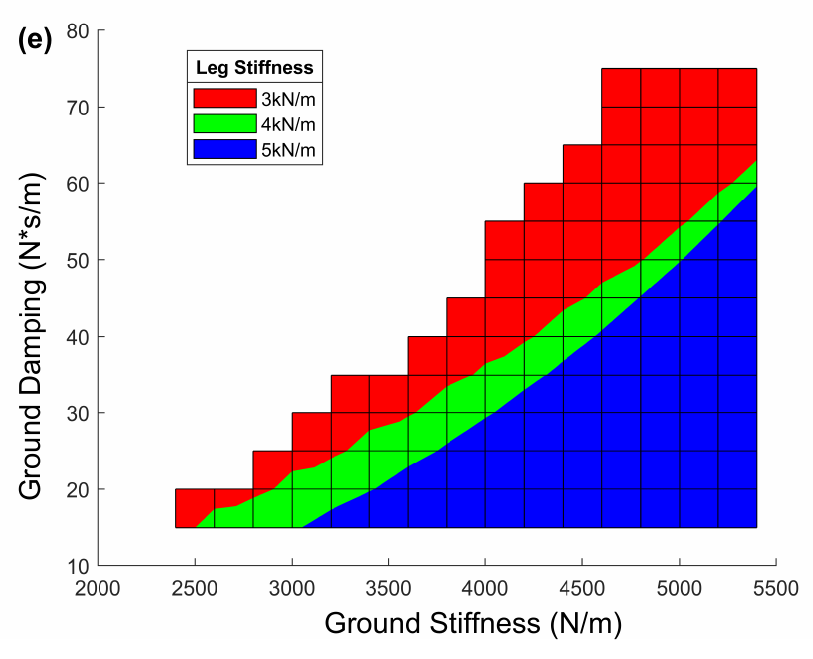}}  
     {\includegraphics[width = 0.24\linewidth]{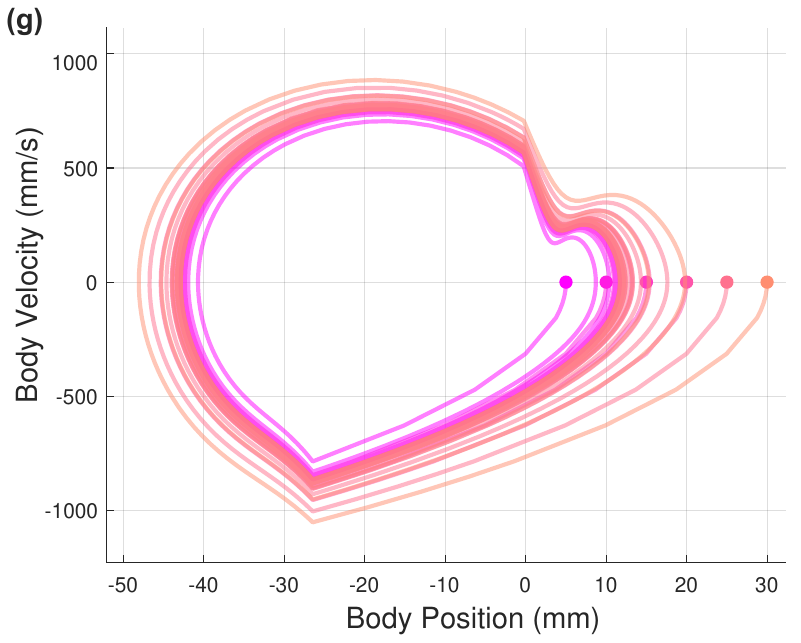}} 

      {\includegraphics[width = 0.24\linewidth]{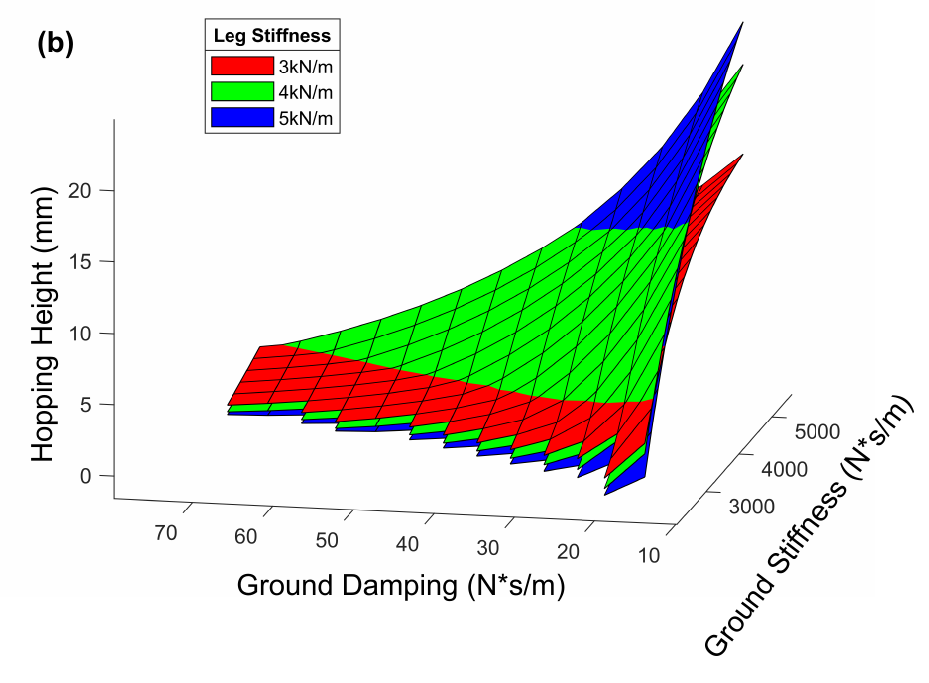}}
    {\includegraphics[width = 0.24\linewidth]{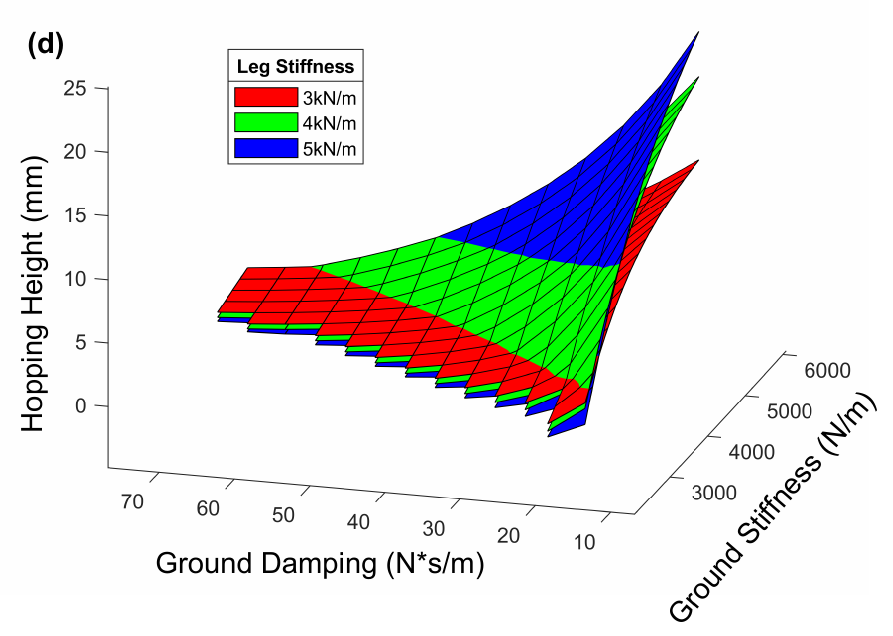}}
      {\includegraphics[width = 0.24\linewidth]{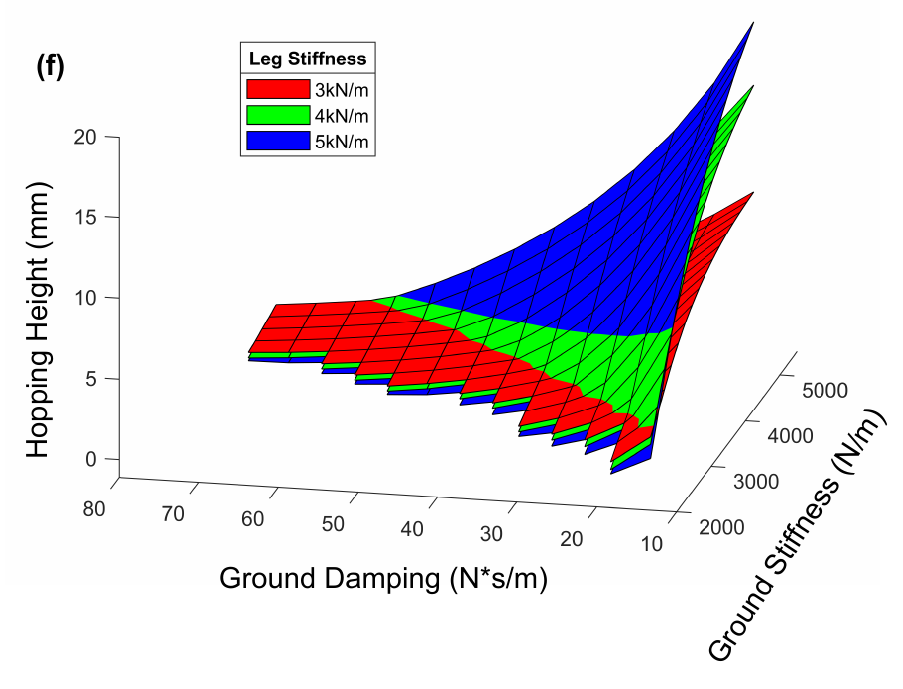}}  
      {\includegraphics[width = 0.24\linewidth]{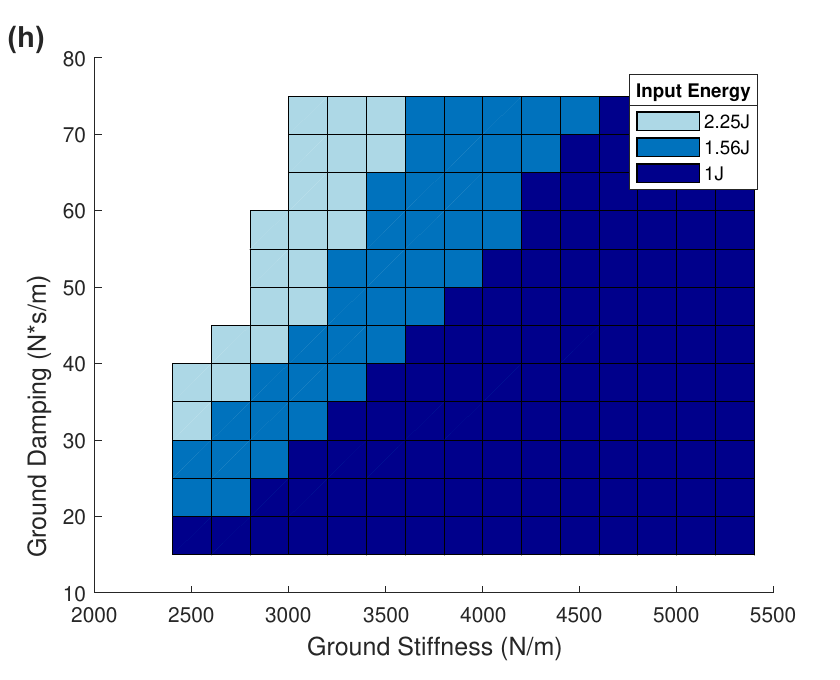}}
      \vspace{-0.5em}
      
     \caption{Hopping height simulation result. Average apex height when (a, b) $d_l=\SI{30}{Ns/m}$, (c, d) $d_l=3\SI{5}{Ns/m}$, and (e, f) $d_l=\SI{40}{Ns/m}$. (g) Simulation of phase portrait ($k_g=\SI{4400}{N/m}$, $k_l = \SI{4300}{N/m}$, $d_g=\SI{35}{Ns/m}, d_l=\SI{35}{Ns/m}$, $E_{in}=\SI{1.5}{J}$) (h) Simulation of successful hopping region.}
     \label{fig: sim result with damping changes}
     \vspace{-0.5em}
 \end{figure*}
The results follow our hypothesis where stiff legs perform well on stiff, less damped ground. 
In fact, the stiff leg performs relatively well in most scenarios, even on more damped ground. 
We will show in simulation that this can be anticipated, as the apex height achieved by the soft legs only slightly outperforms that of the stiff leg, as shown in \figurename~\ref{fig: sim result with damping changes}. 
The soft leg and medium-stiff leg also follow our hypothesis that softer legs perform better on softer, damped ground.
However, there are two unexpected anomalies (discontinued jump that did not follow the trend of the neighboring data points on the grid in \figurename~\ref{fig: exp height figure}) in the ground profiles.
We expected the soft leg to achieve a higher steady-state apex height at $(\SI{2401.7}{N/m}, \SI{35.2}{Ns/m})$, and the medium-stiff leg to perform better at $(\SI{3410.8}{N/m}, \SI{71.4}{Ns/m})$.
We believe this discrepancy is due to unmodeled friction and viscous damping in the vertical rail.
Additionally, video observations showed that the robot exhibited higher lateral oscillation with the soft leg, potentially contributing to energy loss.
Despite these outliers, the remaining data aligns well with the expected trends.



\subsubsection{Simulation Result}

Our simulation takes constant energy as input, along with the robot specifications, leg stiffness, and ground profile, and outputs the steady-state apex height.
\figurename~\ref{fig: sim result with damping changes} (g) shows the phase portrait of the robot's trajectory with parameters closely matching the physical setup: leg stiffness at \SI{4300}{N/m}, ground stiffness at \SI{4400}{N/m}, and both damping coefficients at \SI{35}{Ns/m}.
Each hop is given a fixed input energy of \SI{1.5}{J}, corresponding to a leg compression of \SI{26}{mm}. 
The simulation releases the hopper from six different initial heights, and in all cases, the system converges to a limit cycle.
This example shows that with constant input energy, the hopping height converges to an equilibrium regardless of initial conditions.


The simulated apex heights are shown in \figurename~\ref{fig: sim result with damping changes}, where the colored surfaces represent different leg stiffness inputs.
The data is left blank for trials that failed to achieve steady-state hopping. 
\figurename~\ref{fig: sim result with damping changes} (a-f) highlights the regions where specific leg stiffness values result in the highest apex hopping heights.
Since we did not have a valid way to measure the leg damping coefficient, we repeat the simulation with three sets of different leg damping condition and observed that the results preserved similar trends.

The figure shows four consistent trends: 
(1) Stiffer legs perform better on hard, less damped ground, achieving greater hopping heights, while softer legs excel on softer, more damped surfaces, with stiffer legs experiencing more failures.
(2) Lower ground damping and higher stiffness consistently result in higher jumps, as energy loss is minimized.
(3) When soft legs perform better, the performance difference between leg stiffness is small, but when stiff legs dominate, the difference is more significant. 
(4) As leg damping increases, overall hopping height decreases, and stiff legs gain more advantage.
These results confirm our hypothesis that softer legs are preferable on soft, damped ground to minimize intrusion and energy loss, while stiffer legs perform better on hard, less damped ground by reducing limb deformation and leg damping energy loss. 

Energy input plays a key role in performance. 
We applied three different energy input levels, as listed in Table~\ref{tabHASTA}, with a leg damping coefficient of \SI{35}{Ns/m}, to illustrate where the robot could achieve steady-state hopping. 
These energy values are realistic for real-world scenarios, with the maximum input compressing a \SI{5000}{N/m} leg by \SI{30}{mm}. 
As input energy increased, the ``successful hopping region'' expanded, particularly in softer, more highly damped areas, as shown in \figurename~\ref{fig: sim result with damping changes} (h).

\section{Conclusion}
\label{sec:conclusion}

The stable and consistent vertical hopping performance of HASTA brings us closer to achieving adaptive stiffness control for multi-objective optimization in dynamic locomotion.
Our experimental and simulation results successfully identified the best leg stiffness for various ground profiles, effectively addressing Problem~\ref{prob1}. 
Both methods followed the same trend, confirming our hypothesis: stiffer legs perform better on hard, less damped ground by reducing limb deformation and energy dissipation, while softer legs excel on soft, damped ground by minimizing penetration and energy loss. 
This consistency between experiment and simulation highlights the potential of using the simulation as a guide to selecting the optimal leg stiffness in the controller for improved hopping performance.

Although the stiffer leg performs well across all ground profiles and only marginally loses to the soft leg in certain edge cases, we argue that there are other scenarios where a softer leg would be advantageous. 
For instance, a softer leg is safer to work with and less intrusive on delicate ground surfaces, making it preferable in environmentally sensitive conditions.
Therefore, future work should explore a multi-objective optimization approach that considers additional factors beyond apex height performance, such as specific morphology with point-foot interaction.

With our established methodology, we can extend validation to explore more complex ground models simulated by the ground emulator.
Future work should also focus on developing a comprehensive algorithm to detect different ground profiles in real-time and adjust the leg stiffness accordingly to maximize apex hopping height. 






\bibliographystyle{IEEEtran}
\bibliography{references}

\end{document}